# An Algorithm for Quasi-Associative and Quasi-Markovian Rules of Combination in Information Fusion

Florentin Smarandache
Department of Mathematics and Science
University of New Mexico
200 College Road
Gallup, NM 87301, USA
smarand@unm.edu

Jean Dezert
ONERA/DTIM/IED
29 Avenue de la Division Leclerc
92320 Châtillon, France
Jean.Dezert@onera.fr

**Abstract:** *In this paper one proposes a simple algorithm of combining the fusion rules, those rules which first use the conjunctive rule and then the transfer of conflicting mass to the non-empty sets, in such a way that they gain the property of associativity and fulfill the Markovian requirement for dynamic fusion.*
*Also, a new fusion rule, SDL-improved, is presented.*

**Keywords:** Conjunctive rule, partial and total conflicts, Dempster's rule, Yager's rule, TBM, Dubois-Prade's rule, Dezert-Smarandache classic and hybrid rules, SDL-improved rule, quasi-associative, quasi-Markovian, fusion algorithm

**ACM Classification:** I.2.4.

**1. Introduction.**
We first present the formulas for the conjunctive rule and total conflict, then try to unify some theories using an adequate notation. Afterwards, we propose an easy fusion algorithm in order to transform a quasi-associative rule into an associative rule, and a quasi-Markovian rule into a Markovian rule. One gives examples using the DSm classic and hybrid rules and SDL-improved rule within DSmT. One studies the impact of the VBF on SDLi and one makes a short discussion on the degree of the fusion rules' ad-hoc-ity

**2. The Conjunctive Rule:**
For $n \geq 2$ let $T = \{t_1, t_2, \ldots, t_n\}$ be the frame of discernment of the fusion problem under consideration.
We need to make the remark that in the case when these n elementary hypotheses $t_1, t_2, \ldots, t_n$ are *exhaustive and exclusive* one can use the Dempster-Shafer Theory, Yager's, TBM,

Dubois-Prade Theory, while for the case when the hypotheses are *not exclusive* one can use Dezert-Smarandache Theory, while for non-exhaustivity one uses TBM.

Let m: $2^T \to [0, 1]$ be a basic belief assignment or mass.
The conjunctive rule works in any of these theories, and it is the following in the first theories:

$$\text{for } A \in 2^T, \; m_c(A) = \sum_{\substack{X,Y \in 2^T \\ X \cap Y = A}} m_1(X) m_2(Y) \tag{1}$$

while in DSmT the formula is similar, but instead of the power set $2^T$ one uses the hyper-power set $D^T$, and similarly m: $D^T \to [0, 1]$ be a basic belief assignment or mass:

$$\text{for } A \in D^T, \; m_c(A) = \sum_{\substack{X,Y \in D^T \\ X \cap Y = A}} m_1(X) m_2(Y). \tag{2}$$

The power set is closed under $\cup$, while the hyper-power set is closed under both $\cup$ and $\cap$. Formula (2) allows the use of intersection of sets (for the non-exclusive hypotheses) and it is called DSm classic rule.

The conjunctive rule (1) and its extension (2) to DSmT are associative, which is a nice property needed in fusion combination that we need to extend to other rules derived from it. Unfortunately, only three fusion rules derived from the conjunctive rule are known as associative, i.e. Dempster's rule, Smets's TBM's rule, and Dezert-Smarandache classic rule, the others are not.
For unification of theories let's note by G either $2^T$ or $D^T$ depending on theories.

The conflicting mass $k_{12}$ is computed similarly:
$$k_{12} = m_c(\emptyset) = \sum_{\substack{X,Y \in G \\ X \cap Y = \emptyset}} m_1(X) m_2(Y). \tag{3}$$

Formulas (1), (2), (3) can be generalized for any number of masses $s \geq 2$.

## 3. Associativity.
The propose of this article is to show a simple method to combine the masses in order to keep the associativity and the Markovian requirement, important properties for information fusion.

Let $m_1, m_2, m_3 : G \to [0, 1]$ be any three masses, and a fusion rule denoted by $\oplus$ operating on these masses. One says that this *fusion rule is associative* if:
$$((m_1 \oplus m_2) \oplus m_3)(A) = (m_1 \oplus (m_2 \oplus m_3))(A) \text{ for all } A \in G, \tag{4}$$
which is also equal to $(m_1 \oplus m_2 \oplus m_3)(A)$ for all $A \in G$. \hfill (5)

## 4. Markovian Requirement.

Let $m_1, m_2, \ldots, m_k : G \rightarrow [0, 1]$ be any $k \geq 2$ masses, and a fusion rule denoted by $\oplus$ operating on these masses. One says that this *fusion rule satisfies Markovian requirement* if: $(m_1 \oplus m_2 \oplus \ldots \oplus m_n)(A) = ((m_1 \oplus m_2 \oplus \ldots \oplus m_{n-1}) \oplus m_n)(A)$ for all $A \in G$. (6)

Similarly, only three fusion rules derived from the conjunctive rule are known satisfying the Markovian Requirement, i.e. Dempster's rule, Smets's TBM's rule, and Dezert-Smarandache classic rule.

The below algorithm will help transform a rule into a Markovian rule.

### 5. Fusion Algorithm.
A trivial algorithm is proposed below in order to restore the associativity and Markovian properties to any rule derived from the conjunctive rule.
Let's consider a rule ® formed by using: first the conjunctive rule, noted by ©, and second the transfer of the conflicting mass to non-empty sets, noted by operator "O" (no matter how the transfer is done, either proportionally with some parameters, or transferred to partial or total ignorances and/or to the empty set; if all conflicting mass is transferred to the empty set, as in Smets's rule, there is no need for transformation into an associative or Markovian rule since Smets's rule has already these properties).
Clearly ® = O(©).
The idea is simple, we store the conjunctive rule's result (before doing the transfer) and, when a new mass arises, one combines this new mass with the conjunctive rule's result, not with the result after the transfer of conflicting mass.

Let's have two masses $m_1$, $m_2$ defined as above.
a) One applies the conjunctive rule to $m_1$ and $m_2$ and *one stores* the result:
$m_1 © m_2 = m_{c(1,2)}$ (by notation).
b) One applies the operator O of transferring conflicting mass to the non-empty sets, i.e. $O(m_{c(12)})$.
This calculation completely does the work of our fusion rule, i.e. $m_1 ® m_2 = O(m_{c(12)})$ that we compute for decision-making proposes.
c) When a new mass, $m_3$, arises, we combine using the conjunctive rule this mass $m_3$ with the previous conjunctive rule's result $m_{c(12)}$, not with $O(m_{c(12)})$. Therefore:
$m_{c(1,2)} © m_3 = m_{c(c(1,2),3)}$ (by notation).
One stores this results, while deleting the previous one stored.
d) Now again we apply the operator O to transfer the conflicting mass, i.e. compute $ô(m_{c(c(1,2),3)})$ needed for decision-making.
e) …And so one the algorithm is continued for any number $n \geq 3$ of masses.

The properties of the conjunctive rule, i.e. associativity and satisfaction of the Markovian requirement, are transmitted to the fusion rule ® too.

This is the algorithm we use in DSmT in order to conserve the associativity and Markovian requirement for DSm hybrid rule and SDL improved rule for $n \geq 3$.

Depending on the type of problem to be solved we can use in DSmT either the hybrid rule, or the SDL rule, or a combination of both (i.e., partial conflicting mass is transferred using DSm hybrid, other conflicting mass is transferred using SDL improved rule).

Yet, this easy fusion algorithm can be extended to any rule which is composed from a conjunctive rule first and a transfer of conflicting mass second, returning the associativity and Markovian properties to that rule.

One can remark that the algorithm gives the same result if one applies the rule ⊛ to n ⩾ 3 masses together, and then one does the transfer of conflicting mass.
Within DSmT we designed *fusion rules that can transfer a part of the conflicting mass to partial or total ignorance and the other part of the conflicting mass to non-empty initial sets*, depending on the type of application.

A non-associative rule that can be transformed through this algorithm into an associative rule is called *quasi-associative rule*. And similarly, a non-Markovian rule than can be transformed through this algorithm into a Markovian rule is called *quasi-Markovian rule*.

## 6. SDL-improved Rule.

Let $T = \{t_1, t_2, \ldots, t_n\}$ be the frame of discernment and two masses $m_1, m_2 : G \rightarrow [0, 1]$. One applies the conjunctive rule (1) or (2) depending on theory, then one calculates the conflicting mass (3). In SDL improved rule one transfers partial conflicting masses, instead of the total conflicting mass. If an intersection is empty, say $A \cap B = \emptyset$, then the mass $m(A \cap B)$ is transferred to A and B proportionally with respect to the non-zero sum of masses assigned to A and respectively B by the masses $m_1, m_2$. Similarly, if another intersection, say $A \cap C \cap D = \emptyset$, then again the mass $m(A \cap C \cap D)$ is transferred to A, C, and D proportionally with respect to the non-zero sum of masses assigned to A, C and respectively D by the masses $m_1, m_2$. And so on 'til all conflicting mass is distributed. Then one cumulates the corresponding masses to each non-empty set.

For two masses one has the formula: (7)
$$\text{for } \emptyset \neq A \in D^T, \ m_{\text{SDLi}}(A) = \sum_{\substack{X,Y \in G \\ X \cap Y = A}} m_1(X) m_2(Y) \ + \ c_{12}(A) \cdot \sum_{\substack{X \in G \\ X \cap A = f}} \frac{m_1(X) m_2(A) + m_1(A) m_2(X)}{c_{12}(A) + c_{12}(X)}$$

where $c_{12}(A)$ is the non-zero sum of the mass matrix column corresponding to the set A, i.e.
$$c_{12}(A) = m_1(A) + m_2(A) \neq 0. \tag{8}$$

For more masses one applies the algorithm to formulas (7) and (8).

## 7. Ad-Hoc-ity of Fusion Rules.
Each fusion rule is more or less ad-hoc. Same thing for SDL improved. There is up to the present no rule that fully satisfies everybody. Let's analyze some of them.
*Dempster's rule* transfers the conflicting mass to non-empty sets proportionally with their resulting masses. What is the reasoning for doing this? Just to swallow the masses of non-empty sets in order to sum up to 1?

*Smets's rule* transfers the conflicting mass to the empty set. Why? Because, he says, we consider on open world where unknown hypotheses might be. Not convincing.

*Yager's rule* transfers the conflicting mass to the total ignorance. Should the conflicting mass be ignored?

*Dubois-Prade's rule* and DSm hybrid rule transfers the conflicting mass to the partial and total ignorances. Not completely justified either.

*SDL improved rule* is based on partial conflicting masses, transferred to the corresponding sets proportionally with respect to the non-zero sums of their assigned masses. But other weighting coefficients can be found. Inagaki (1991), Lefevre-Colot-Vannoorenberghe (2002) proved that there are infinitely many fusion rules based on the conjunctive rule and then on the transfer of the conflicting mass, all of them depending on the weighting coefficients that transfer that conflicting mass. How to choose them, what parameters should they rely on – that's the question! There is not a measure for this.

In my opinion, neither DSm hybrid rule nor SDLi rule are not more ad-hoc than other fusion rules.

"No matter how you do, people will have objections" (Wu Li).

**8. Numerical Examples**.

We show how it is possible to use the above fusion algorithm in order to transform a quasi-associative and quasi-Markovian rule into an associative and Markovian one.

Let T = {A, B, C}, all hypotheses exclusive, and two masses $m_1$, $m_2$ that form the corresponding mass matrix:

|       | A   | B   | A∪C |
|-------|-----|-----|-----|
| $m_1$ | 0.4 | 0.5 | 0.1 |
| $m_2$ | 0.6 | 0.2 | 0.2 |

8.1 Let's take the DSm hybride rule:

8.1.1. Let's check the associativity:

a) First we use the DSm classic rule and we get at time t1:

$m_{DSmC12}(A)=0.38$, $m_{DSmC12}(B)=0.10$, $m_{DSmC12}(A∪C)=0.02$, $m_{DSmC12}(A∩B)=0.38$,

$m_{DSmC12}(B∩(A∪C))=0.12$, and one stores this result. (S1)

b) One uses the DSm hybrid rule and we get:

$m_{DSmH12}(A)=0.38$, $m_{DSmH12}(B)=0.10$, $m_{DSmH12}(A∪C)=0.02$, $m_{DSmH12}(A∪B)=0.38$,

$m_{DSmH12}(A∪B∪C)=0.12$. This result was computed because it is needed for decision making on two sources/masses only. (R1)

c) A new masses, $m_3$, arise at time t2, and has to be taken into consideration, where $m_3(A)=0.7$, $m_3(B)=0.2$, $m_3(A∪C)=0.1$.

Now one combines the result stored at (S1) with $m_3$, using DSm classic rule, and we get:

$m_{DSmC(12)3}(A)=0.318$, $m_{DSmC(12)3}(B)=0.020$, $m_{DSmC(12)3}(A∪C)=0.002$,

$m_{DSmC(12)3}(A∩B)=0.610$, $m_{DSmC(12)3}(B∩(A∪C))=0.050$, and one stores this result, (S2) while deleting (S1).

d) One uses the DSm hybrid rule and we get:

$m_{DSmH(12)3}(A)=0.318$, $m_{DSmH(12)3}(B)=0.020$, $m_{DSmH(12)3}(A∪C)=0.002$,

$m_{DSmH(12)3}(A∪B)=0.610$, $m_{DSmH(12)3}(A∪B∪C)=0.050$. This result was also computed because it is needed for decision making on three sources/masses only. (R2)

e) And so on for as many masses as needed.

First combining the last masses, $m_2$, $m_3$, one gets:
$m_{DSmC23}(A)=0.62$, $m_{DSmC23}(B)=0.04$, $m_{DSmC23}(A\cup C)=0.02$, $m_{DSmC23}(A\cap B)=0.26$,
$m_{DSmC23}(B\cap (A\cup C))=0.06$, and one stores this result. (S3)
Using DSm hybrid one gets:
$m_{DSmH23}(A)=0.62$, $m_{DSmH23}(B)=0.04$, $m_{DSmH23}(A\cup C)=0.02$, $m_{DSmH23}(A\cup B)=0.26$,
$m_{DSmH23}(A\cup B\cup C)=0.06$.
Then, combining $m_1$ with $m_{DSmC23}$ {stored at (S3)} using DSm classic and then using DSm hybrid one obtain the same result (R2).

If one applies the DSm hybride rule to all three masses together one gets the same result (R2).

We showed on this example that <u>DS</u>m hybrid applied within the <u>a</u>lgorithm is associative (i.e. using the notation DSmHa one has):
DSmHa(($m_1$, $m_2$), $m_3$) = DSmHa($m_1$, ($m_2$, $m_3$)) = DSmHa($m_1$, $m_2$, $m_3$).

8.1.2. Let's check the Markov requirement:
a) Combining three masses together using DSm classic:

|       | A   | B   | A∪C |       |
|-------|-----|-----|-----|-------|
| $m_1$ | 0.4 | 0.5 | 0.1 | (M1)  |
| $m_2$ | 0.6 | 0.2 | 0.2 |       |
| $m_3$ | 0.7 | 0.2 | 0.1 |       |

one gets as before:
$m_{DSmC123}(A)=0.318$, $m_{DSmC123}(B)=0.020$, $m_{DSmC123}(A\cup C)=0.002$, $m_{DSmC123}(A\cap B)=0.610$,
$m_{DSmC123}(B\cap (A\cup C))=0.050$, and one stores this result in (S2).
b) One uses the DSm hybrid rule to transfer the conflicting mass and we get:
$m_{DSmH123}(A)=0.318$, $m_{DSmH123}(B)=0.020$, $m_{DSmH123}(A\cup C)=0.002$, $m_{DSmH123}(A\cup B)=0.610$,
$m_{DSmH123}(A\cup B\cup C)=0.050$.
c) Suppose a new mass $m_4$ arises, $m_4(A)=0.5$, $m_4(B)=0.5$, $m_4(A\cup C)=0$.
Use DSm classic to combine $m_4$ with $m_{DSmC123}$ and one gets:
$m_{DSmC(123)4}(A)=0.160$, $m_{DSmC(123)4}(B)=0.010$, $m_{DSmC(123)4}(A\cup C)=0$, $m_{DSmC(123)4}(A\cap B)=0.804$,
$m_{DSmC(123)4}(B\cap (A\cup C))=0.026$, and one stores this result in (S3).
d) Use DSm hybrid rule:
$m_{DSmH(123)4}(A)=0.160$, $m_{DSmH(123)4}(B)=0.010$, $m_{DSmH(123)4}(A\cup C)=0$, $m_{DSmH(123)4}(A\cup B)=0.804$,
$m_{DSmH(123)4}(A\cup B\cup C)=0.026$. (R4)

Now, if one combines all previous four masses, $m_1$, $m_2$, $m_3$, $m_4$, together using first the DSm classic then the DSm hybrid one still get (R4). Whence the Markovian requirement.
We didn't take into account any discounting of masses.

8.2. Let's use the SDL improved rule on the same example.
a) One considers the above mass matrix (M1) and one combines $m_1$ and $m_2$ using DSm classic and one gets as before:
$m_{DSmC12}(A)=0.38$, $m_{DSmC12}(B)=0.10$, $m_{DSmC12}(A\cup C)=0.02$, $m_{DSmC12}(A\cap B)=0.38$,
$m_{DSmC12}(B\cap (A\cup C))=0.12$, and one stores this result in (S1).

b) One transfers the partial conflicting mass 0.38 to A and B respectively:
$x/1 = y/0.7 = 0.38/1.8$; whence $x=0.223529$, $y=0.156471$.
One transfers the other conflicting mass 0.12 to B and A∪C respectively:
$z/0.7 = w/0.3 = 0.12/1$; whence $z=0.084$, $w=0.036$.
One cumulates them to the corresponding sets and one gets:
$m_{SDLi12}(A) = 0.38+0.223529 = 0.603529$;
$m_{SDLi12}(B) = 0.10+0.156471 + 0.084 = 0.340471$;
$m_{SDLi12}(A∪C) = 0.02+0.036 = 0.056000$.
c) One uses the DSm classic rule to combine the above $m_3$ and the result in (S1) and one gets again:
$m_{DSmC(12)3}(A)=0.318$, $m_{DSmC(12)3}(B)=0.020$, $m_{DSmC(12)3}(A∪C)=0.002$,
$m_{DSmC(12)3}(A∩B)=0.610$, $m_{DSmC(12)3}(B∩(A∪C))=0.050$, and one stores this result in (S2) while deleting (S1).
d) One transfers the partial conflicting masses 0.610 to A and B respectively, and 0.050 to B and A∪C respectively. Then one cumulates the corresponding masses and one gets:
$m_{SDLi(12)3}(A) = 0.716846$;
$m_{SDLi(12)3}(B) = 0.265769$;
$m_{SDLi(12)3}(A∪C) = 0.017385$.

Same result we obtain if one combine first $m_2$ and $m_3$, and the result combine with $m_1$, or if we combine all three masses $m_1$, $m_2$, $m_3$ together.

## 9. Vacuous Belief Function.

SDLi seems to satisfy Smets's impact of VBF (Vacuum Belief Function. i.e. m(T)=1), because there is no partial conflict ever between the total ignorance T and any of the sets of G. Since in SDLi the transfer is done after each partial conflict, T will receive no mass, not being involved in any partial conflict. Thus VBF acts as a neutral elements with respect with the composition of masses using SDLi. The end combination does not depend on the number of VBF's included in the combination.
Let's check this on the previous example. Considering the first two masses $m_1$ and $m_2$ in (M1) and using SDLi one got: $m_{SDLi12}(A) = 0.603529$, $m_{SDLi12}(B) = 0.340471$, $m_{SDLi12}(A∪C) = 0.056000$.
Now let's combine the VBF too:

|     | A   | B   | A∪C | A∪B∪C |       |
|-----|-----|-----|-----|-------|-------|
| VBF | 0   | 0   | 0   | 1     | (M2)  |
| $m_1$ | 0.4 | 0.5 | 0.1 | 0     |       |
| $m_2$ | 0.6 | 0.2 | 0.2 | 0     |       |

a) One uses the DSm classic rule to combine all three of them and one gets again:
$m_{DSmC(VBF12)}(A)=0.38$, $m_{DSmC(VBF12)}(B)=0.10$, $m_{DSmC(VBF12)}(A∪C)=0.02$,
$m_{DSmC(VBF12)}(A∩B)=0.38$, $m_{DSmC(VBF12)}(B∩(A∪C))=0.12$, $m_{DSmC(VBF12)}(A∪B∪C)=0$ and one stores this result in (S1).
b) One transfers the partial conflicting mass 0.38 to A and B respectively:
$x/1 = y/0.7 = 0.38/1.8$; whence $x=0.223529$, $y=0.156471$.
One transfers the other conflicting mass 0.12 to B and A∪C respectively:
$z/0.7 = w/0.3 = 0.12/1$; whence $z=0.084$, $w=0.036$.

Therefore nothing is transferred to the mass of A∪B∪C, then the results is the same as above: $m_{SDLi12}(A) = 0.603529$, $m_{SDLi12}(B) = 0.340471$, $m_{SDLi12}(A\cup C) = 0.056000$.

## 10. Conclusion.

We propose an elementary fusion algorithm that transforms any fusion rule (which first uses the conjunctive rule and then the transfer of conflicting masses to non-empty sets, except for Smets's rule) to an associative and Markovian rule. This is very important in information fusion since the order of combination of masses should not matter, and for the Markovian requirement the algorithm allows the storage of information of all previous masses into the last result (therefore not necessarily to store all the masses), which later will be combined with the new mass.

In DSmT, using this fusion algorithm for $n \geqslant 3$ sources, the DSm hybrid rule and SDLi are commutative, associative, Markovian, and SDLi also satisfies the impact of vacuous belief function.